\documentclass[letterpaper, 10 pt, conference]{ieeeconf}  

\IEEEoverridecommandlockouts                              

\usepackage{graphics} 
\usepackage{graphicx}
\usepackage{graphicx}
\usepackage{epsfig} 
\usepackage{mathptmx} 
\usepackage{times} 
\usepackage{amsmath} 
\usepackage{amssymb}  
\usepackage{amsmath}
\usepackage{booktabs}
\usepackage{microtype}
\usepackage{makecell}  
\usepackage{booktabs}  

\allowdisplaybreaks

\usepackage{amsmath,amsfonts}
\usepackage{algorithmic}
\usepackage{algorithm}
\usepackage{float}
\title{\LARGE \bf
Dynamic Modeling and Dimensional Optimization of Legged Mechanisms for Construction Robots}

\author{Xiao Liu$^{1}$, Xianlong Yang$^{1,2}$, Weijun Wang$^{1,3}$, Wei Feng $^{1,3*}$
\thanks{*corresponding author. e-mail: wei.feng@siat.ac.cn}
\thanks{$^{1}$ All authors are with Shenzhen Institute of Advanced Technology, Chinese Academy of Sciences, Shenzhen, 518055, China.
        {\tt\small  Contact: xiao.liu1@siat.ac.cn}}%
 \thanks{$^{2}$ All authors are with Wuhan University of Technology, Wuhan, 430070, China.
        {\tt\small  }}%
 \thanks{$^{3}$ All authors are with the University of Chinese Academy of Sciences, Shenzhen, 518055, China.
        {\tt\small  }}%
}

\begin{document}

\maketitle
\thispagestyle{empty}
\pagestyle{empty}

\begin{abstract}

With the rapid development of the construction industry, issues such as harsh working environments, high-intensity and high-risk tasks, and labor shortages have become increasingly prominent. This drives higher demands for construction robots in terms of low energy consumption, high mobility, and high load capacity. This paper focuses on the design and optimization of leg structures for construction robots, aiming to improve their dynamic performance, reduce energy consumption, and enhance load-bearing capabilities. Firstly, based on the leg configuration of ants in nature, we design a structure for the robot's leg. Secondly, we propose a novel structural optimization method. Using the Lagrangian approach, a dynamic model of the leg was established. Combining the dynamic model with the leg's motion trajectory, we formulated multiple dynamic evaluation metrics and conducted a comprehensive optimization study on the geometric parameters of each leg segment. The results show that the optimized leg structure reduces peak joint torques and energy consumption by over 20\%. Finally, dynamic simulation experiments were conducted using ADAMS. The results demonstrate a significant reduction in the driving power of each joint after optimization, validating the effectiveness and rationality of the proposed strategy. This study provides a theoretical foundation and technical support for the design of heavy-load, high-performance construction robots.

 This paper is currently under review at Mechanics Based Design of Structures and Machines.

\end{abstract}

\section{INTRODUCTION}

With the accelerating advancement of global urbanization, the construction industry faces increasingly severe challenges [1]. Construction sites commonly suffer from harsh environments, high-intensity workloads, and significant safety hazards, which directly contribute to a persistently high rate of accidents in the industry [2]. Construction robots reduce accident rates significantly by replacing humans in performing hazardous tasks, heavy lifting, and repetitive installations. However, the application of construction robots still encounters several challenges, the most prominent being high energy consumption, low load capacity, and high costs [3]. Particularly when handling heavy construction materials such as ceilings, prefabricated walls, and large tiles, robots must overcome greater energy consumption and torque demands, placing higher requirements on their overall performance and economic feasibility [4]. To address these issues, we design a hexapod chassis for construction robots, which enhances load-bearing capacity and adaptability to complex terrains, enabling it as a legged robot.
\begin{figure}[htb]
      \centering
      \includegraphics[width=\columnwidth]{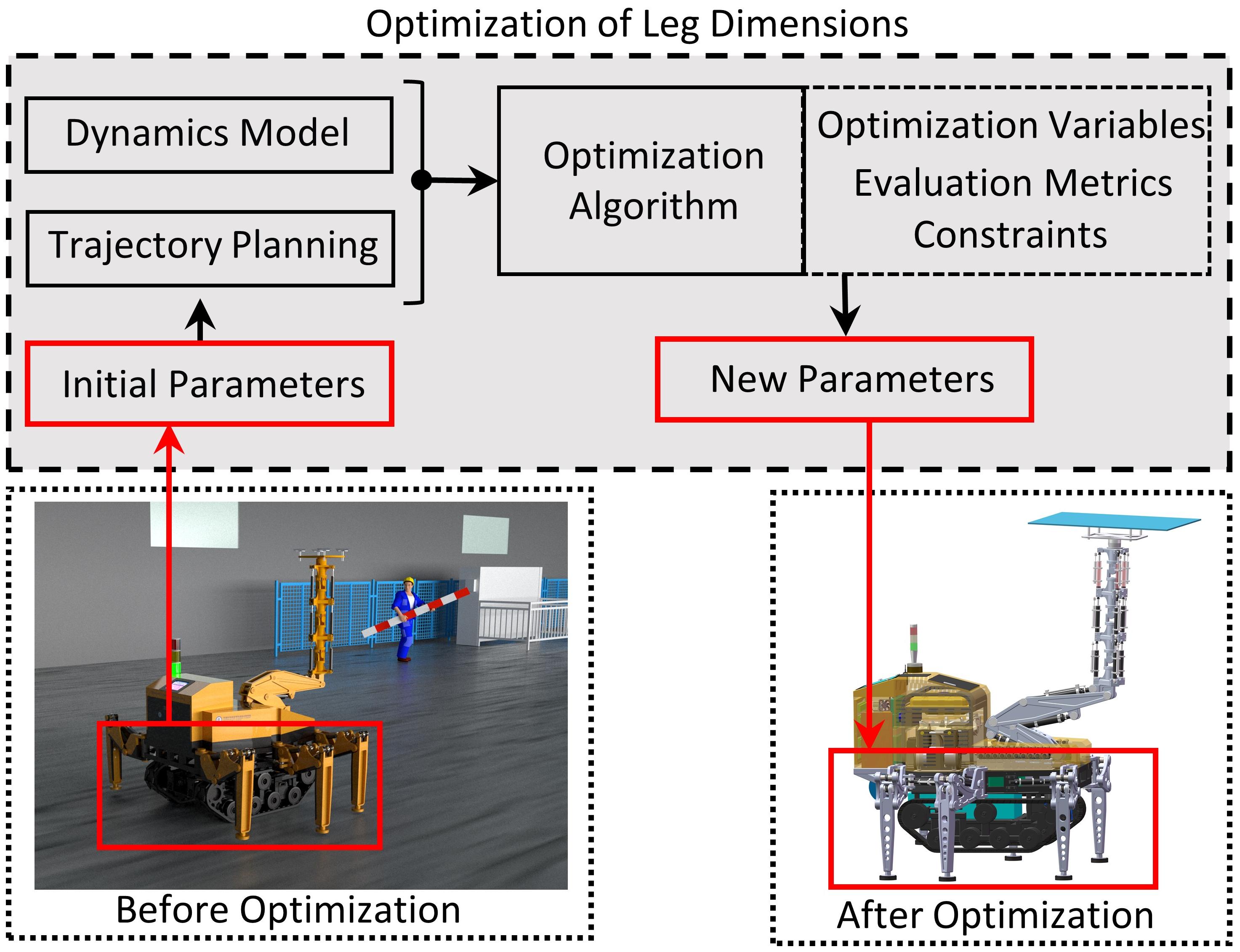}
      \caption{Optimization of Leg Dimensions}
      \label{figurelabel}
   \end{figure}

Regarding legged robots, many researchers have adopted various methods to optimize their structures [5-12]. Among these, [13] proposes a design method for heavy-duty robotic legs based on force generation mechanisms for high-load robots. This work derives a dimensionless EMA expression for the robot leg and incorporates it into structural optimization. [14] uses the workspace as a constraint and the main structural dimensions as design variables, optimizing the size of a 3-TPT parallel mechanism to enhance stiffness within the workspace. [15] designs a novel multi-modal motion robot named SHOALBOT, performing optimization and performance evaluation. Through optimization, the thrust of the propeller legs in water is increased by 400\%, and the minimum difference between forward and reverse thrust is optimized from 25\% to 3\%. [16] proposes a wearable medical robot and obtains optimal parameters for hip and knee joint variables using particle swarm optimization (PSO). [17] conducts design optimization using three key performance criteria: global condition number, link forces norm, and robot compliance, followed by multi-criteria optimization using an evolutionary algorithm. [18] develops a computational method for leg kinematics and integrates it into an optimization algorithm with defined boundary conditions. After optimization, the stride length of the walking leg increases by 29.23\%, and the influence of swim leg geometric parameters on swimming performance is analyzed. [19] establishes a structural size optimization model for a single leg, demonstrating that the optimized leg can meet obstacle-crossing performance metrics.

The aforementioned researchers optimize the structure and dimensions of robotic legs to improve task performance, driving force, and motion trajectory accuracy. However, they do not simultaneously address the technical challenges of high load capacity, high energy consumption, and high mobility required for construction robots. Inspired by the leg configurations of multi-legged organisms in nature, we propose a preliminary design for the overall leg configuration and single-leg structure of construction robots to enhance mobility and load-bearing capacity. In addition to length, the various parts of the robot's legs have geometric parameters such as width and height, which determine the mass and moment of inertia of each leg segment, thereby influencing the leg's dynamic performance. Assuming the construction robot moves at a constant speed, the leg undergoes complex motion states during its swing phase, including acceleration during lifting and deceleration during lowering, accompanied by conversions between kinetic and potential energy. The geometric parameters significantly affect the dynamic performance of the leg during the swing phase. Therefore, based on the leg's swing operation, this paper establishes multiple dynamic evaluation metrics and conducts a comprehensive optimization study of the geometric parameters of each leg segment. This approach aims to address the challenges of high load capacity and high energy consumption in heavy-duty construction robots.

\section{STRUCTURAL DESIGN OF THE LEG}
Inspired by the high load-bearing capacity and stability of ants in nature, a biomimetic hexapod mobile chassis is applied to construction robots to meet the demands of carrying large loads and complex terrains in construction environments. Currently, the leg design of hexapod biomimetic robots most commonly adopts a three-degree-of-freedom configuration. The structure of these three-degree-of-freedom legs is primarily categorized into articulated, Cartesian, and telescopic types, as shown in Figure 2.
\begin{figure}[htb]
      \centering
      \includegraphics[width=\columnwidth]{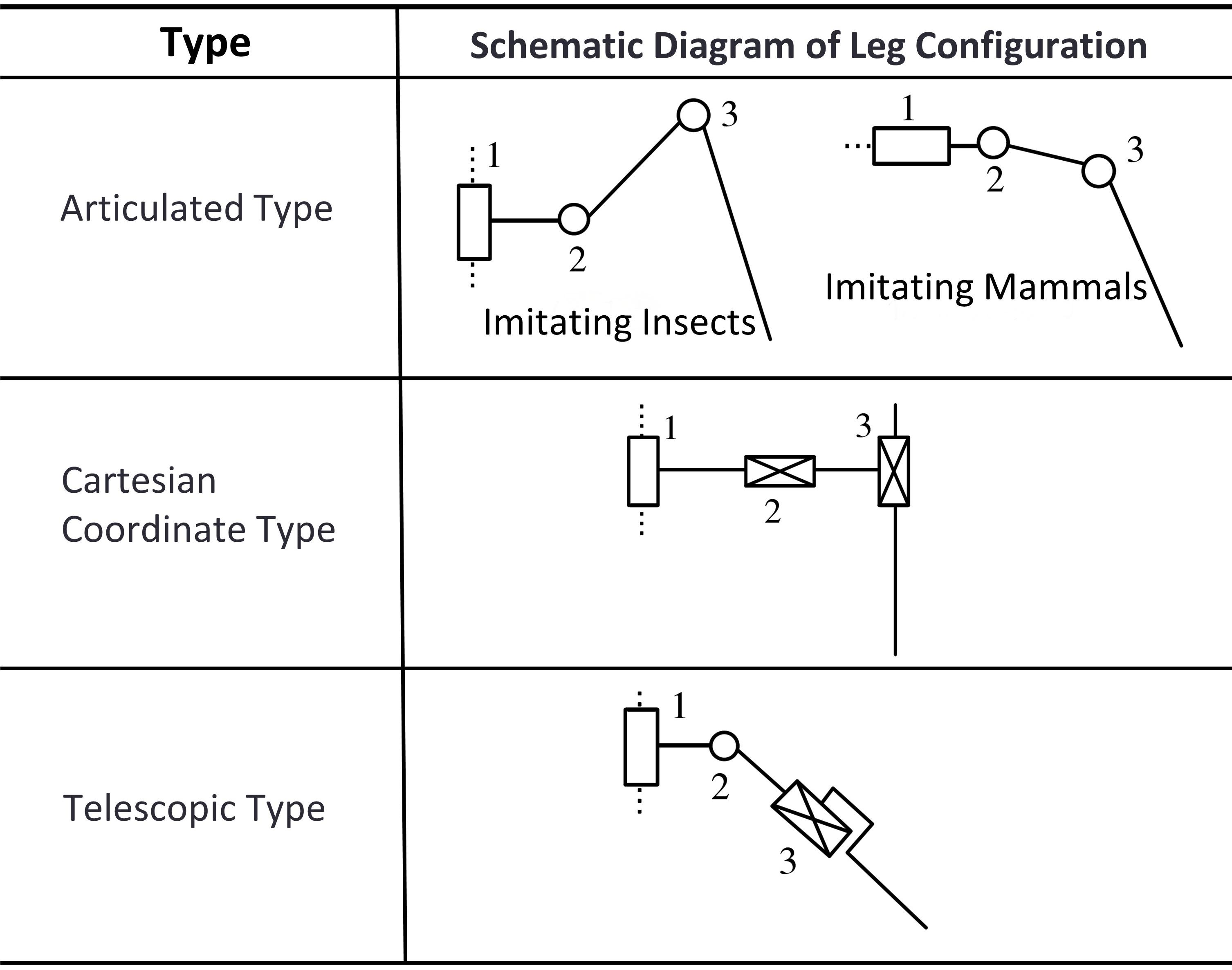}
      \caption{Typical Configurations of Three-Degree-of-Freedom Legs}
      \label{figurelabel}
   \end{figure}

Since this construction robot mimics ants to achieve high load-bearing capacity, an articulated leg structure is adopted, as shown in Figure 3. The single leg mainly consists of a coxa, femur, and tibia, which are connected by three rotational joints: the root joint (Joint 1), hip joint (Joint 2), and knee joint (Joint 3).
\begin{figure}[htb]
      \centering
      \includegraphics[width=\columnwidth]{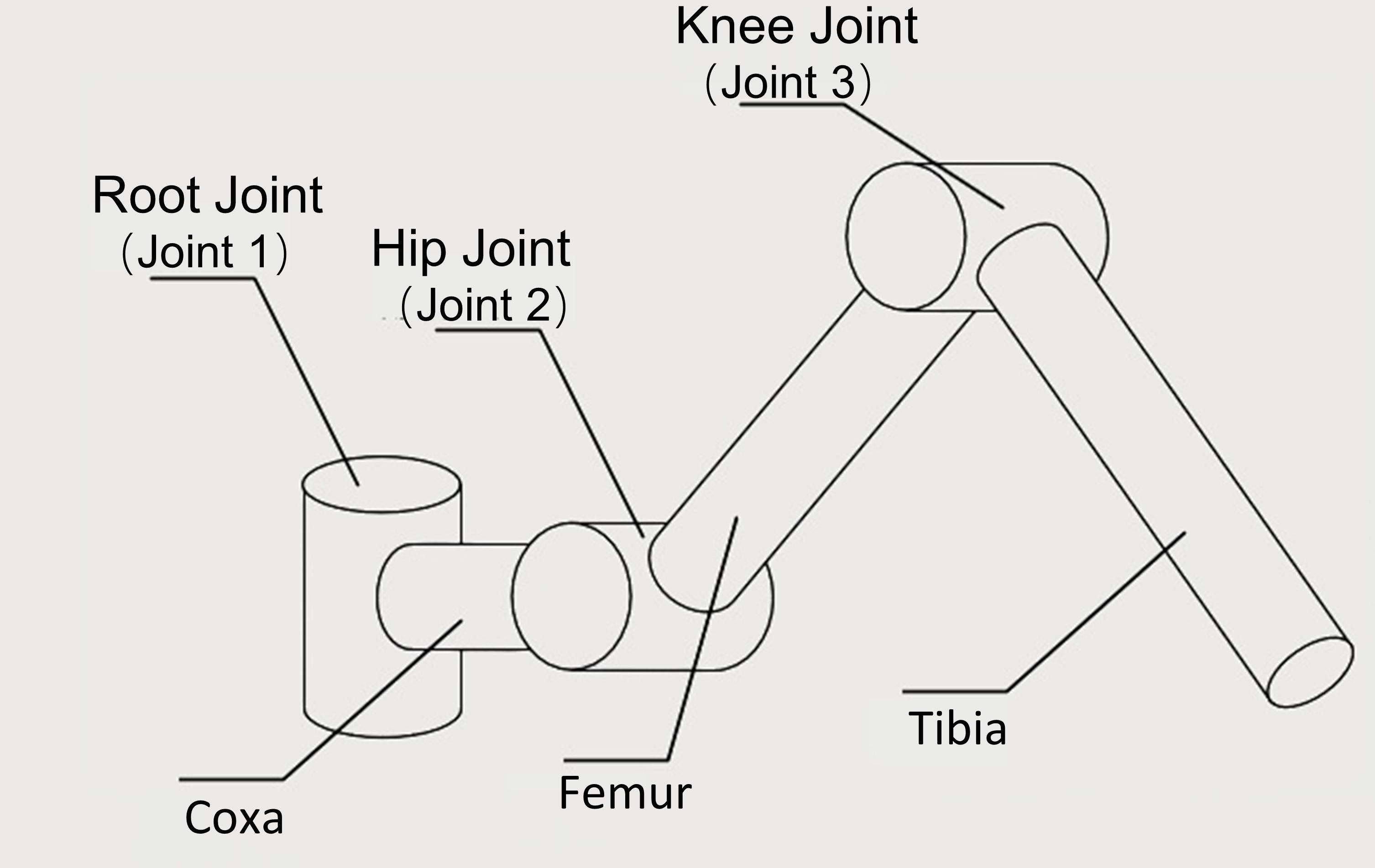}
      \caption{ Leg Configuration of the Construction Robot}
      \label{figurelabel}
   \end{figure}

For legs with the same configuration, the dimensional proportions between leg segments and the size parameters of each structure significantly impact the dynamic performance of hexapod robots. Therefore, structural optimization of the legs is of great importance.
\section{TRAJECTORY PLANNING OF THE LEG}
Based on the characteristics of leg analysis, trajectory planning in joint space is adopted. Since construction robots need to overcome obstacles of a certain height in construction environments, intermediate points, specifically the leg-lifting positions, must be considered in the leg trajectory planning. Although high-order polynomials can satisfy more constraints, an increase in the order may lead to deviations from the ideal trajectory curve and raise computational complexity. Therefore, this paper employs quintic polynomials for multi-segment path planning and applies smoothing techniques to transition through the intermediate points.

Let each segment of the path have a starting point \( A \) and an endpoint \( B \), with the start time of the motion being \( t_A \)  and the end time being \( t_B \). For trajectory planning based on joint space, each segment of the motion trajectory has the following six constraints: the initial joint angle \( \theta_A \), angular velocity \( \dot{\theta}_A \), and angular acceleration \( \ddot{\theta}_A \); the final joint angle \( \theta_B \), angular velocity \( \dot{\theta}_B \), and angular acceleration \( \ddot{\theta}_B \). In joint space-based trajectory planning for the legs, the position, velocity, and acceleration at specified positions are represented by joint angles, angular velocities, and angular accelerations. The root joint, hip joint, and knee joint angles of the construction robot's leg at the start point, midpoint, and endpoint are respectively $[-\pi/4, -\pi/4, -\pi/4]$, $[0, \pi/4, -3\pi/4]$, and $[\pi/4, -\pi/4, -\pi/4]$. The angular velocities at the start point, midpoint, and endpoint are respectively $[0, 0, 0]$, $[\pi/2, 0, 0]$, and $[0, 0, 0]$. The angular accelerations at the start point, midpoint, and endpoint are respectively $[0, 0, 0]$, $[0, 0, 0]$, and $[0, 0, 0]$. The total motion time is 2 seconds, with the time from the start point to the midpoint and from the midpoint to the endpoint both being 1 second. Based on quintic polynomial planning, when the leg moves from the start point to the midpoint, the functions for the changes in joint angles, angular velocities, and angular accelerations over time are as follows:

Root joint (joint 1):
\begin{equation}
\label{deqn_ex1}
\begin{cases}
\theta(t)=-\frac{\pi}{4}+\frac{\pi}{2}t^{3}-\frac{\pi}{4}t^{4} \\
\dot{\theta}(t)=\frac{3\pi}{2}t^{2}-\pi t^{3} \\
\ddot{\theta}(t)=3\pi t-3\pi t^{2} & 
\end{cases}
\end{equation}
Hip Joint (Joint 2):
\begin{equation}
\label{deqn_ex1}
\begin{cases}
\theta(t)=-\frac{\pi}{4}+5\pi t^3-\frac{15}{2}\pi t^4+3\pi t^5 \\
\dot{\theta}(t)=15\pi t^2-30\pi t^3+15\pi t^4 \\
\ddot{\theta}(t)=30\pi t-90\pi t^2+60\pi t^3 & 
\end{cases}
\end{equation}
Knee joint (joint 3):
\begin{equation}
\label{deqn_ex1}
\begin{cases}
\theta(t)=-\frac{\pi}{4}-5\pi t^3+\frac{15\pi}{2}t^4-3\pi t^5 \\
\dot{\theta}(t)=-15\pi t^2+30\pi t^3-15\pi t^4 \\
\ddot{\theta}(t)=-30\pi t+90\pi t^2-60\pi t^3 & 
\end{cases}
\end{equation}
Similarly, the functions for the changes in joint angles, angular velocities, and angular accelerations over time as the leg moves from the midpoint to the endpoint are calculated as follows:

Root joint (joint 1):
\begin{equation}
\label{deqn_ex1}
\begin{cases}
\theta(t)=\frac{\pi}{2}t-\frac{\pi}{2}t^3+\frac{\pi}{4}t^4 \\
\dot{\theta}(t)=\frac{\pi}{2}-\frac{3\pi}{2}t^2+\pi t^3 \\
\ddot{\theta}(t)=-3\pi t+3\pi t^2 & 
\end{cases}
\end{equation}
Hip Joint (Joint 2):
\begin{equation}
\label{deqn_ex1}
\begin{cases}
\theta(t)=\frac{\pi}{4}-5\pi t^3+\frac{15\pi}{2}t^4-3\pi t^5 \\
\dot{\theta}(t)=-15\pi t^2+30\pi t^3-15\pi t^4 \\
\ddot{\theta}(t)=-30\pi t+90\pi t^2-60\pi t^3 & 
\end{cases}
\end{equation}
Knee joint (joint 3):
\begin{equation}
\label{deqn_ex1}
\begin{cases}
\theta(t)=-\frac{3\pi}{4}+5\pi t^3-\frac{15}{2}\pi t^4+3\pi t^5 \\
\dot{\theta}(t)=15\pi t^2-30\pi t^3+15\pi t^4 \\
\ddot{\theta}(t)=30\pi t-90\pi t^2+60\pi t^3 & 
\end{cases}
\end{equation}
To observe the specific trajectory curves, set the initial lengths of the coxa, femur, and tibia as \( l_1 = 200 \, \text{mm} \), \( l_2 = 400 \, \text{mm} \), and \( l_3 = 400 \, \text{mm} \). Based on the changes in joint angles from the start point to the midpoint and from the midpoint to the endpoint, use MATLAB combined with forward kinematics to calculate and plot the end-effector trajectory curve, as shown in Figure 4.
\begin{figure}[htb]
      \centering
      \includegraphics[width=\columnwidth]{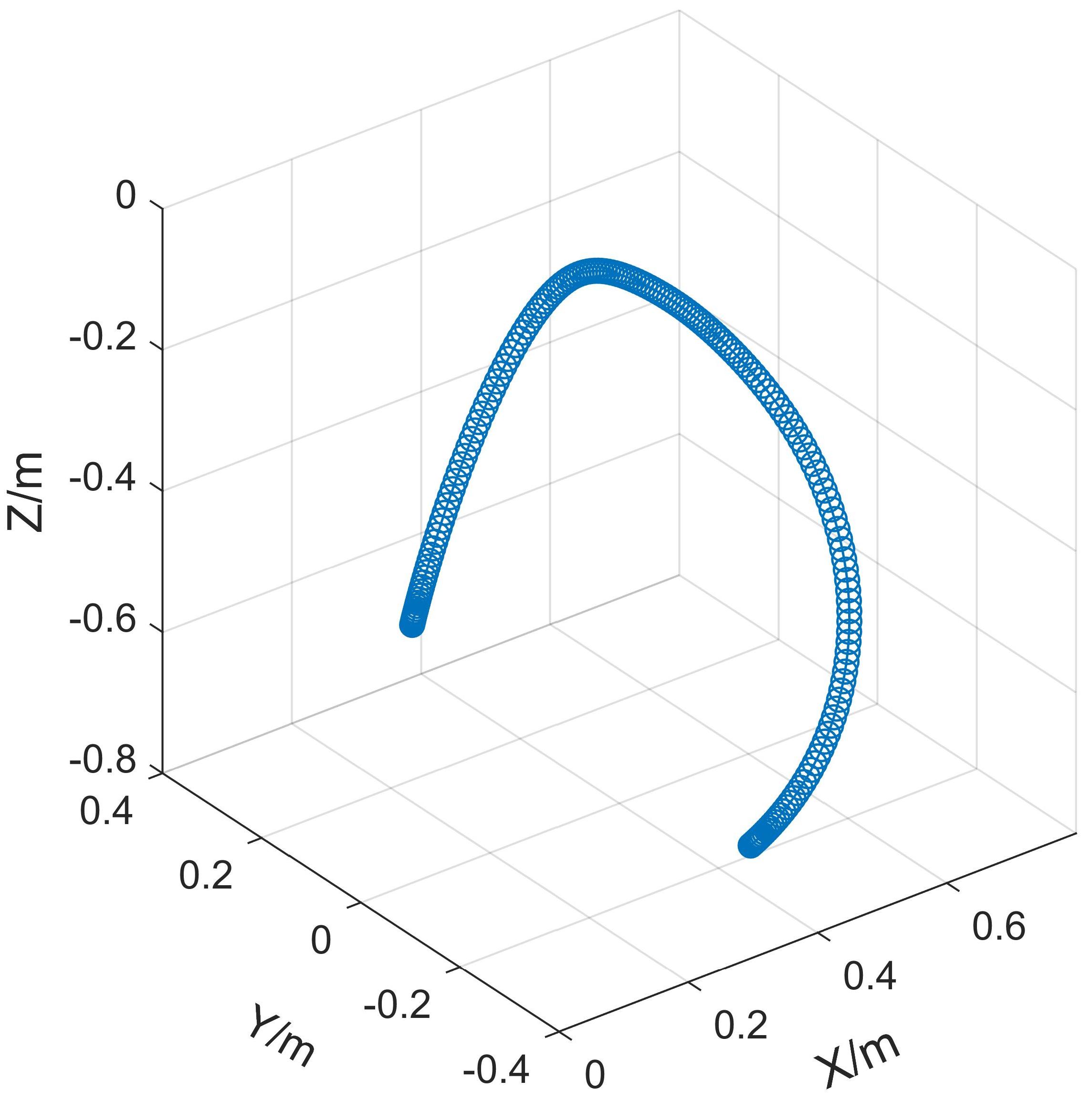}
      \caption{Trajectory of the Leg Endpoint}
      \label{figurelabel}
   \end{figure}
   
As can be seen from the above figure, the planned trajectory curve is smooth without abrupt changes, demonstrating the reasonableness of the trajectory planning. Based on Equations (1) to (6), the curves of joint angles, angular velocities, and angular accelerations at different times during the leg's motion along the planned trajectory are plotted, as shown in Figure 5.
\begin{figure}[htb]
      \centering
      \includegraphics[width=\columnwidth]{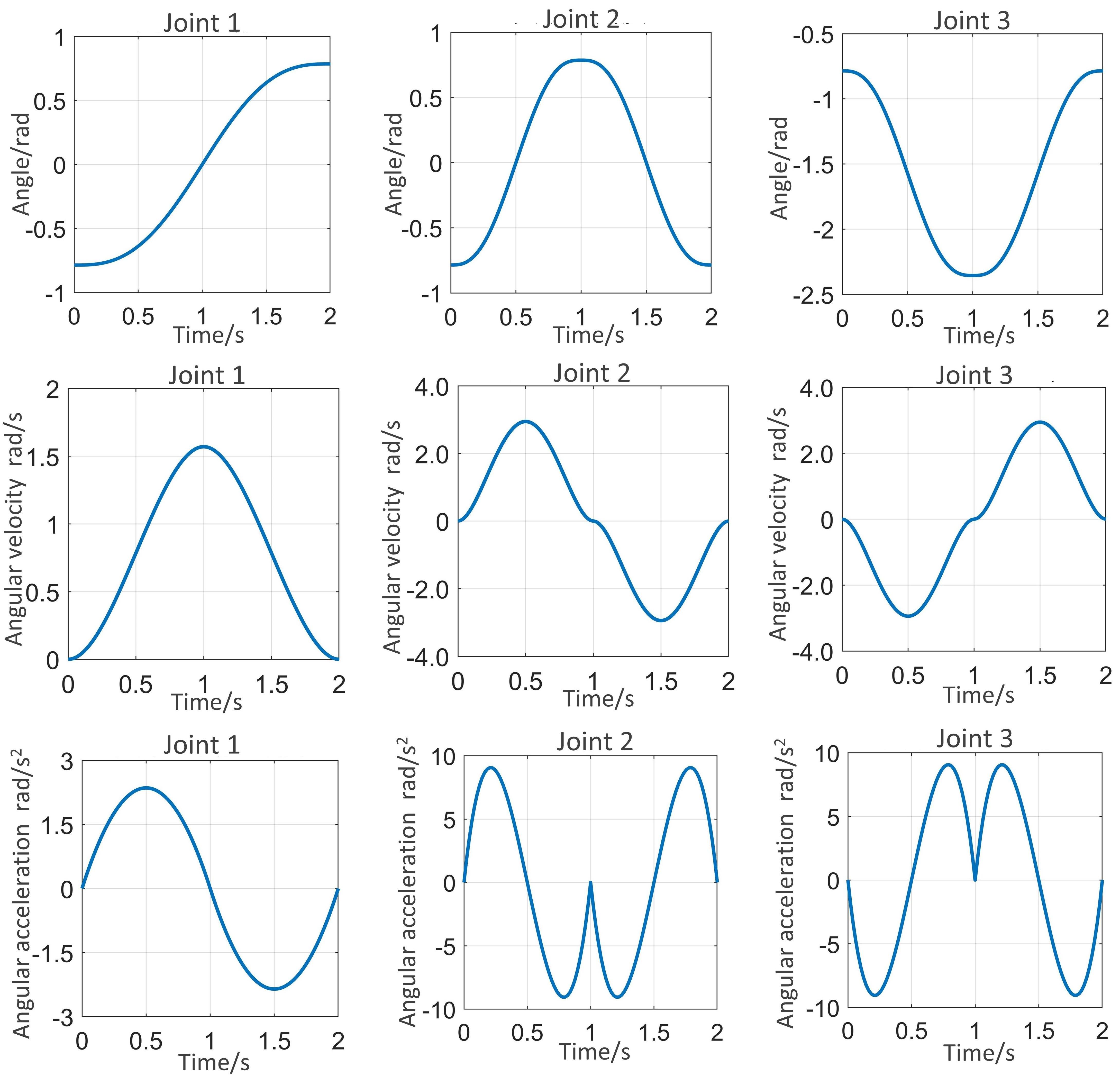}
      \caption{Angular Acceleration Variation Curves of Joints}
      \label{figurelabel}
   \end{figure}

It can be observed that when the leg moves along the planned trajectory, the joint angles of the root joint \(\theta_1\), hip joint \(\theta_2\), and knee joint \(\theta_3\) do not exhibit any abrupt changes. Additionally, the angular velocity and angular acceleration curves are smooth and continuous, proving that the leg motion is stable and the trajectory planning is reasonable.
\section{DYNAMIC MODELING}
To study the relationships between the joint angles, angular velocities, angular accelerations, and the dimensions of each leg segment with the torques at each joint, a dynamic model of the leg is required. Compared to the Newton-Euler approach, using Lagrange's method allows for a more intuitive understanding of the relationship between external forces and moments on each link and the motion of the links. Therefore, this paper employs Lagrange's equations for the dynamic modeling of the leg. The Lagrangian function \( L \) is defined as the difference between the total kinetic energy \( E_k \) and the total potential energy \( E_p \):
\begin{equation}
\label{deqn_ex1}
L=E_k-E_p
\end{equation}
Thus, the Lagrangian equation of motion is derived as follows:
\begin{equation}
\label{deqn_ex1}
F_{i}=\frac{d}{dt}\frac{\partial L}{\partial\dot{x}_{i}}-\frac{\partial L}{\partial x_{i}}
\end{equation}
Among these, \( F_i \) represents the sum of all external forces that produce motion, \( x_i \) represents the variables in the system.

For a leg in motion, the potential energy of each part of the leg depends only on the angular position \( \theta \) and not on the angular velocity \( \dot{\theta} \). Therefore, Equation (7) can be expressed as:
\begin{equation}
\label{deqn_ex1}
L\left(\theta,\dot{\theta}\right)=E_k\left(\theta,\dot{\theta}\right)-E_p\left(\theta\right)
\end{equation}
The leg can be considered a three-axis robotic arm, where \( \tau \) represents the sum of all external torques that produce rotation, i.e., the torques at each joint. The variable is the joint angle \( \theta \). Therefore, for each leg segment, the Lagrangian equation of motion is:
\begin{equation}
\label{deqn_ex1}
\tau=\frac{d}{dt}\frac{\partial L}{\partial\dot{\theta}}-\frac{\partial L}{\partial\theta}
\end{equation}
Substituting Equation (10) into the above equation yields:
\begin{equation}
\label{deqn_ex1}
\tau=\frac{d}{dt}\frac{\partial E_k}{\partial\dot{\theta}}-\frac{\partial E_k}{\partial\theta}+\frac{\partial E_p}{\partial\theta}
\end{equation}
Equation (11) is the Lagrangian equation of motion for each leg segment.
From Equation (9), the Lagrangian function \( L \) is the difference between the total kinetic energy \( E_k \) and the total potential energy \( E_p \) of the system. Therefore, the kinetic and potential energies of each leg segment need to be determined.

(1) Determine the kinetic energy \( E_k \) of each leg segment.
Let \( r_i \) be an arbitrary point on the local coordinate system of the \( i \)-th link. The position of this point in the base coordinate system can be determined using the transformation matrix \( {}^0T_i \) derived from the kinematic analysis. The velocity of this point is:
\begin{equation}
\label{deqn_ex1}
\nu_i=\frac{d}{dt}\left(^0T_ir_i\right)=\sum_{j=1}^i\left(\frac{\partial^0T_i}{\partial\theta_j}\dot{\theta}_j\right)\cdot r_i
\end{equation}
Then, for the mass element \(m_i\) located at the point \(r_i\), its kinetic energy is:
\begin{equation}
\label{deqn_ex1}
dK_i=\frac{1}{2}\left(\dot{x}_i^2+\dot{y}_i^2+\dot{z}_i^2\right)dm_i=\frac{1}{2}Trace\left(\nu_i\nu_i^T\right)dm_i
\end{equation}
Combining Equations (12) and (13) yields the kinetic energy of point \( r_i \) as:
\begin{equation}
\resizebox{1\hsize}{!}{$
\label{deqn_ex1}
dK_i=\frac{1}{2}Trace\left[\left(\sum_{r=1}^i\left(\frac{\partial^0T_i}{\partial\theta_r}\dot{\theta}_r\right)\cdotp r_i\right)\left(\sum_{k=1}^i\left(\frac{\partial^0T_i}{\partial\theta_k}\dot{\theta}_k\right)\cdotp r_i\right)^T\right]dm_i
$}
\end{equation}
Therefore, for the \( i \)-th link, the kinetic energy of translation is:
\begin{equation}
\resizebox{1\hsize}{!}{$
\label{deqn_ex1}
\begin{aligned}
K_{1i} & 
\begin{aligned}
=\int dK_i=\frac{1}{2}Trace\left[\sum_{r=1}^i\sum_{k=1}^i\frac{\partial^0T_i}{\partial\theta_r}\left(\int r_ir_i^Tdm_i\right)\left(\frac{\partial^0T_i}{\partial\theta_k}\right)^T\dot{\theta}_r\dot{\theta}_k\right]
\end{aligned} \\
 & =\frac{1}{2}Trace{\left[\sum_{r=1}^{i}\sum_{k=1}^{i}\frac{\partial^{0}T_{i}}{\partial\theta_{r}}J_{i}\left(\frac{\partial^{0}T_{i}}{\partial\theta_{k}}\right)^{T}\dot{\theta}_{r}\dot{\theta}_{k}\right]}
\end{aligned}
$}
\end{equation}
Where \( J_i \) is the pseudo-inertia matrix, and \( r \) and \( k \) denote the indices of different joints. The rotational kinetic energy of the \( i \)-th link can be easily expressed as:
\begin{equation}
\label{deqn_ex1}
K_{2i}=\frac{1}{2}I_i\dot{\theta}_i^2
\end{equation}
Where \( I_i \) is the rotational inertia of each link relative to its joint axis.
Therefore, the total kinetic energy \( K_i \) of the \( i \)-th link is the sum of the translational kinetic energy \( K_{1i} \) and the rotational kinetic energy \( K_{2i} \), i.e.:
\begin{equation}
\label{deqn_ex1}
K_i=K_{1i}+K_{2i}
\end{equation}
Substituting equations (15) and (16) into equation (17), and assuming that each leg segment is a regular-shaped rod with uniformly distributed mass, the total kinetic energy of each leg segment can be derived as follows:

Coxa:
\begin{equation}
\label{deqn_ex1}
E_{k1}=\frac{1}{2}m_1a_1^2\dot{\theta}_1^2+\frac{1}{2}I_1\dot{\theta}_1^2
\end{equation}
Femur:
\begin{equation}
\label{deqn_ex1}
E_{k2}=\frac{1}{2}m_2\left[\left(l_1+\alpha_2\cos\theta_2\right)^2\dot{\theta}_1^2+\alpha_2^2\dot{\theta}_2^2\right]+\frac{1}{2}I_2\dot{\theta}_2^2
\end{equation}
Tibia:
\begin{equation}
\label{deqn_ex1}
\begin{aligned}
E_{k3} & = \frac{1}{2}m_3 \Biggl[ (l_1 + l_2\cos\theta_2 + a_3\cos(\theta_3 - \theta_2))^2 \dot{\theta}_1^2 + 2l_2a_3\dot{\theta}_2 (\dot{\theta}_2\\ 
& \quad - \dot{\theta}_3)\cos\theta_3+ l_2^2\dot{\theta}_2^2 + \alpha_3^2\left(\dot{\theta}_2 - \dot{\theta}_3\right)^2 \Biggr] + \frac{1}{2}I_3\dot{\theta}_3^2
\end{aligned}
\end{equation}

Where \( m_i \) is the mass of each link, and \( a_i \) is the distance from the center of mass of each link to the joint axis.
(2) Determine the potential energy \( E_p \) of each leg segment.
The total potential energy \( P_i \) of the \( i \)-th link is:
\begin{equation}
\label{deqn_ex1}
P_i=-m_ig^T\left({}^0T_i\overline{r_i}\right)
\end{equation}
Where \( \mathbf{g}^T \) represents the gravity matrix, and \( \vec{r}_i \) represents the position of the center of mass of the link in the link coordinate system.

The total potential energy of each leg segment can be obtained from Equation (21):

Coxa:
\begin{equation}
\label{deqn_ex1}
E_{p1}=m_1g_1h
\end{equation}

Femur:
\begin{equation}
\label{deqn_ex1}
E_{p2}=m_2g\left(h+a_2\sin\theta_2\right)
\end{equation}

Tibia:
\begin{equation}
\label{deqn_ex1}
E_{p3}=m_3g\left(h+l_2\sin\theta_2-a_3\sin\left(\theta_3-\theta_2\right)\right)
\end{equation}
Where \( h \) is the distance from the center of mass of the base link to the zero potential energy plane.
Since each leg segment is considered a regular-shaped, uniform rod, \( a_i = l_i / 2 \). From Equations (18) to (20) and Equations (22) to (24), we obtain:

For the root joint:
\begin{equation}
  \resizebox{1\hsize}{!}{$
\label{deqn_ex1}
\begin{aligned}
\frac d{dt}\frac{\partial E_k}{\partial\dot{\theta}_1} & 
\begin{aligned}
= & \left[\frac{1}{2}m_1l_1^2+m_2\left(l_1+\frac{1}{2}l_2\cos\theta_2\right)^2+m_3\left(l_1+l_2\cos\theta_2+\frac{1}{2}l_2\cos(\theta_3-\theta_2)\right)^2+I_1\right]\ddot{\theta}_1
\end{aligned} \\
 & -
\begin{bmatrix}
m_2l_2\sin\theta_2\bigg(l_1+\frac12l_2\cos\theta_2\bigg) \\
 \\
+2m_3\bigg(l_1+l_2\cos\theta_2+\frac12l_3\cos(\theta_3-\theta_2)\bigg)\bigg(l_2\sin\theta_2-\frac12l_3\cos(\theta_3-\theta_2)\bigg)\bigg]
\end{bmatrix}\dot{\theta}_1\dot{\theta}_2 \\
 & -m_3l_3\sin\left(\theta_3-\theta_2\right){\left(l_1+l_2\cos\theta_2+\frac12l_3\cos(\theta_3-\theta_2)\right)}\dot{\theta}_1\dot{\theta}_3
\end{aligned}
$}
\end{equation}

\begin{equation}
\label{deqn_ex1}
\frac{\partial E_k}{\partial\theta_1}=0
\end{equation}

\begin{equation}
\label{deqn_ex1}
\frac{\partial E_p}{\partial\theta_1}=0
\end{equation}
For the hip joint:

\begin{equation}
\label{deqn_ex1}
\begin{aligned}
\frac d{dt}\frac{\partial E_k}{\partial\dot{\theta}_2} =& \left[\frac{1}{4}m_2l_2^2+m_3\left(l_2^2+l_2l_3\cos\theta_3+\frac{1}{4}l_3^2\right)+I_2\right]\ddot{\theta}_2\\
& -m_3\left(\frac{1}{4}l_3^2+\frac{1}{2}l_2l_3\cos\theta_3\right)\ddot{\theta}_3 \\
& +\frac{1}{2}m_3l_2l_3\sin\theta_3\dot{\theta}_3^2-m_3l_2l_3\sin\theta_3\dot{\theta}_2\dot{\theta}_3
\end{aligned}
\end{equation}

\begin{equation}
\label{deqn_ex1}
\begin{aligned}
\frac{\partial E_k}{\partial\theta_2}& =-\Bigg\lfloor m_{3}\Bigg(l_{1}+l_{2}\cos\theta_{2}+\frac{1}{2}l_{3}\cos\big(\theta_{3}-\theta_{2}\big)\Bigg)\Bigg(l_{2}\sin\theta_{2}-\\
&\frac{1}{2}l_{3}\sin\big(\theta_{3}-\theta_{2}\big)\Bigg) +\frac12m_2l_2\sin\theta_2\biggl(\frac12l_2\cos\theta_2+l_1\biggr)\biggr]\dot{\theta}_1^2
\end{aligned}
\end{equation}

\begin{equation}
\label{deqn_ex1}
\frac{\partial E_p}{\partial\theta_2}=\frac{1}{2}m_2gl_2\cos\theta_2+m_3gl_2\cos\theta_2+\frac{1}{2}m_3gl_3\cos(\theta_3-\theta_2)
\end{equation}

For the knee joint:
\begin{equation}
\label{deqn_ex1}
\begin{split}
\frac{d}{dt}\frac{\partial E_k}{\partial\dot{\theta}_3}=&(\frac{1}{4}m_3l_3^2+I_3)\ddot{\theta}_3-m_3(\frac{1}{4}l_3^2+\frac{1}{2}l_2l_3\cos\theta_3)\ddot{\theta}_2 \\
&+\frac{1}{2}m_3l_2l_3\sin\theta_3\dot{\theta}_2\dot{\theta}_3
\end{split}
\end{equation}

\begin{equation}
\label{deqn_ex1}
\begin{split}
\frac{\partial E_k}{\partial\theta_3} =& -\frac{1}{2}m_3 \Biggl[l_3\sin(\theta_3 - \theta_2) (l_1 + l_2\cos\theta_2 + \frac{1}{2}l_3\cos(\theta_3 \\
& -\theta_2))\dot{\theta}_1^2+ l_2l_3\sin\theta_3\dot{\theta}_2^2 - l_2l_3\sin\theta_3\dot{\theta}_2\dot{\theta}_3 \Biggr]
\end{split}
\end{equation}

\begin{equation}
\label{deqn_ex1}
\frac{\partial E_p}{\partial\theta_3}=-\frac{1}{2}m_3gl_3\cos(\theta_3-\theta_2)
\end{equation}
Substituting the above expressions into the Lagrangian equation of motion (Eq. 11) yields the dynamic equations for each leg segment, representing the relationship between joint torques and the structural parameters of the leg segments.

\section{DYNAMICS-BASED DIMENSIONAL OPTIMIZATION}

\subsection{Optimization Variables}
The robot's leg consists of geometric parameters such as length, width, and height. The optimization targets are the geometric dimensions of the leg. To facilitate the setup of optimization variables and dynamic calculations, the coxa, femur, and tibia segments of the leg are assumed to be uniform rods with hollow rectangular cross-sections. The simplification ensures that the mass, center of mass, and moment of inertia of each leg segment remain consistent before and after the simplification. A schematic diagram of the simplified leg model and its dimensional parameters is shown in Fig.6.
\begin{figure}[htb]
      \centering
      \includegraphics[width=\columnwidth]{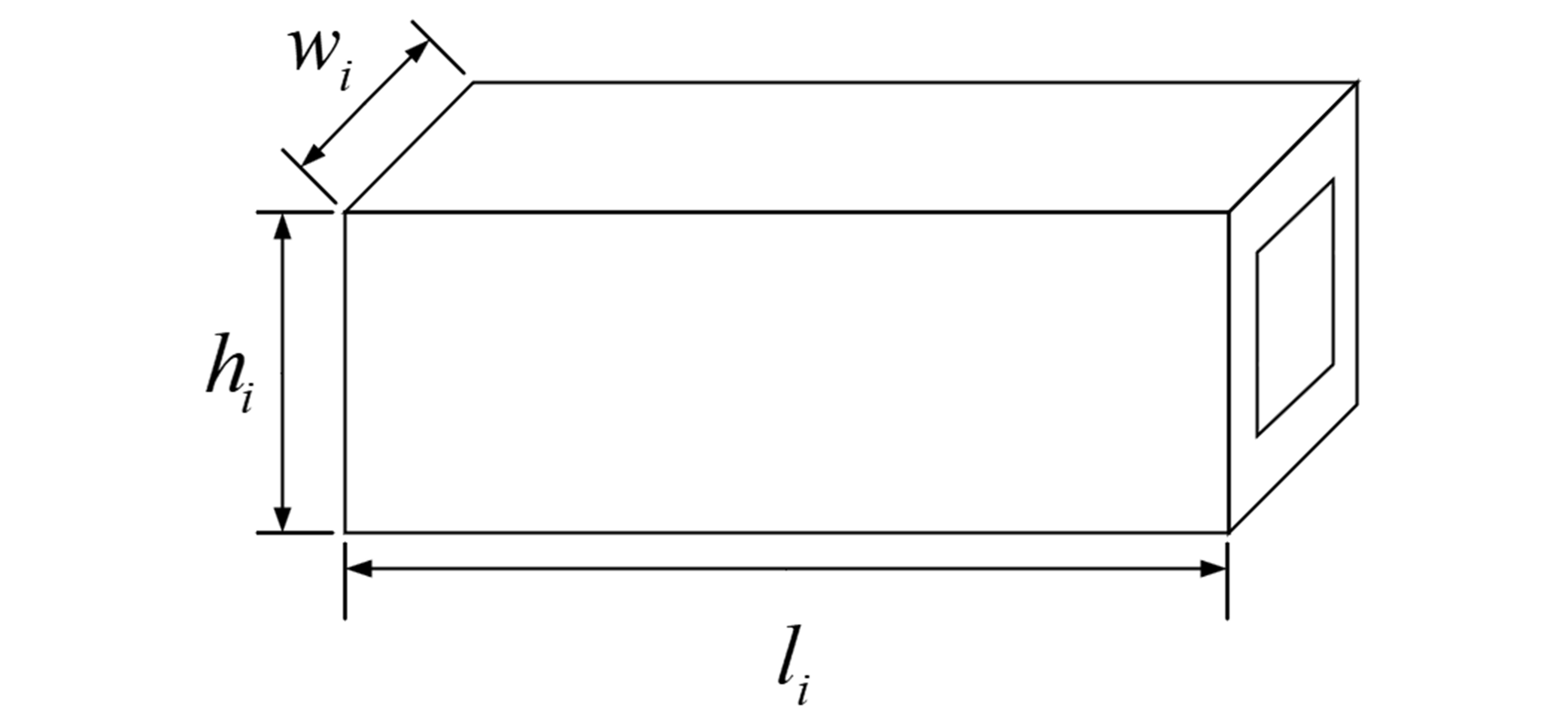}
      \caption{Schematic Diagram of the Simplified Leg Segment Model}
      \label{Fig.6}
   \end{figure}

The optimization variables are defined as the length, width, and height of each leg segment, i.e., \( l_1, l_2, l_3, w_1, w_2, w_3, h_1, h_2, h_3 \), totaling nine variables. The initial structural parameters of each leg segment can be set as shown in Table 1.

\begin{table}[ht]
    \centering
    \caption{Initial Structural Parameters of the Leg}
    \resizebox{8.5cm}{!}{%
       \begin{tabular}{cccc}
       \toprule
       \textbf{Leg Segment} & 
       \textbf{\makecell{Rod Length \\ $ l $ (mm)}} &  
       \textbf{\makecell{Cross-Sectional\\ Dimensions \\ $ h \times w $ (mm)}} & 
       \textbf{\makecell{Mass \\ $ m $ (kg)}} \\       
       \midrule
       Coxa & 140 & $ 179 \times 121 $ & 6.06 \\
       Femur & 460 & $ 158 \times 183 $ & 20.09 \\
       Tibia & 460 & $ 144 \times 117 $ & 14.22 \\
       \bottomrule
      \end{tabular}%
    }  
\end{table}

Based on stablished dynamics models of each leg segment, the relationship between the joint torques and the geometric dimensions, mass, and angular parameters (angle \( \theta \), angular velocity \( \dot{\theta} \), and angular acceleration \( \ddot{\theta} \)) of each joint can be obtained. By importing the discrete trajectory points into MATLAB and substituting them into the established dynamics equations, the variation curves of the joint torques under the initial structural parameters can be plotted, as shown in Fig.7.
\begin{figure}[htb]
      \centering
      \includegraphics[width=\columnwidth]{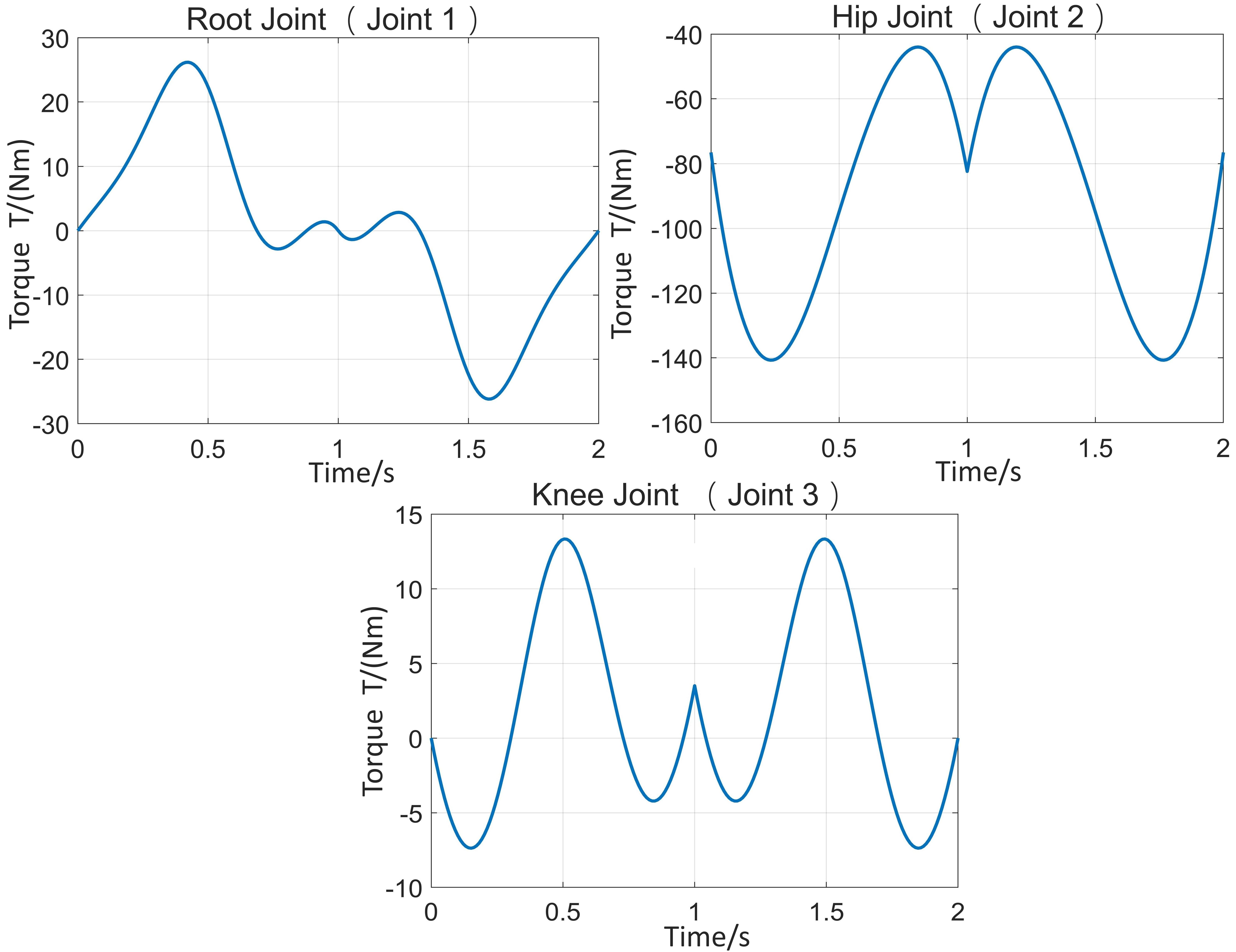}
      \caption{Torque Variation Curves of the Joints}
      \label{Fig.7}
   \end{figure}

It can be observed that when the leg moves along the planned trajectory, the torques at the root joint and knee joint are relatively small, while the torque at the hip joint is significantly larger. Based on an analysis of the motion trajectory, the following observations are made regarding the leg's movement: (1) The torque at the root joint is primarily related to the kinetic energy of the coxa, femur, and tibia segments rotating around the root joint axis. Consequently, the root joint torque remains relatively small. (2) The torque at the hip joint is mainly associated with the kinetic energy of the femur and tibia segments rotating around the hip joint axis, as well as the gravitational potential energy of the femur and tibia. As a result, the hip joint experiences a larger torque. (3) The torque at the knee joint is primarily influenced by the kinetic energy of the tibia segment rotating around the knee joint axis and the gravitational potential energy of the tibia. However, due to the relatively small change in gravitational potential energy of the tibia relative to the femur, the knee joint torque remains relatively small.
\subsection{Evaluation Metrics}
From Figure 7, it is evident that each joint torque of the leg reaches a peak during the motion trajectory. Excessive peak torques can damage the structure and drive system. Thus, the peak joint torque is defined as the first evaluation metric for optimizing the leg structure, denoted as \( T_M \), and expressed as:
\begin{equation}
\label{deqn_ex1}
T_{Mi}=\max\left\{\left|\tau_i(\theta)\right|\right\}
\end{equation}
Compared to conventional industrial robots, construction robots require high mobility for field operations and cannot rely on a continuous power supply. Therefore, reducing energy consumption to enhance operational endurance is a critical consideration in structural optimization. Joint energy consumption is thus defined as the second evaluation metric for structural optimization, denoted as \( Q \), and expressed as:
\begin{equation}
\label{deqn_ex1}
Q_i=\sum_{j=1}^NP_i^jT_i^j=\sum_{j=1}^N\tau_i^j\dot{\theta}_i^jT_i^j
\end{equation}

In the equation, \( N \) is the number of discrete characteristic points on the leg's motion trajectory; \( P_i^j \) is the power of the \( i \)-th joint at the \( j \)-th point on the trajectory; \( \tau_i^j \) is the torque of the \( i \)-th joint; \( \theta_i^j \) is the angular velocity of the \( i \)-th joint; and \( T_i^j \) is the time interval for the leg to move from the \( (j-1) \)-th point to the \( j \)-th point, where \( T_i^j = t_j - t_{j-1} \).
\subsection{Optimization Algorithm}

As discussed above, there are nine variables to be optimized and two evaluation metrics, making it difficult to determine the optimal solution using traditional computational methods. Therefore, this paper employs a genetic algorithm to solve the multi-variable optimization problem. A genetic algorithm
generally consists of key components such as encoding, initial population, fitness function, genetic operators, and termination conditions. The specific optimization process of the algorithm is illustrated in Fig.8.
   \begin{figure}[htb]
    \centering
    \includegraphics[width=0.7\columnwidth]{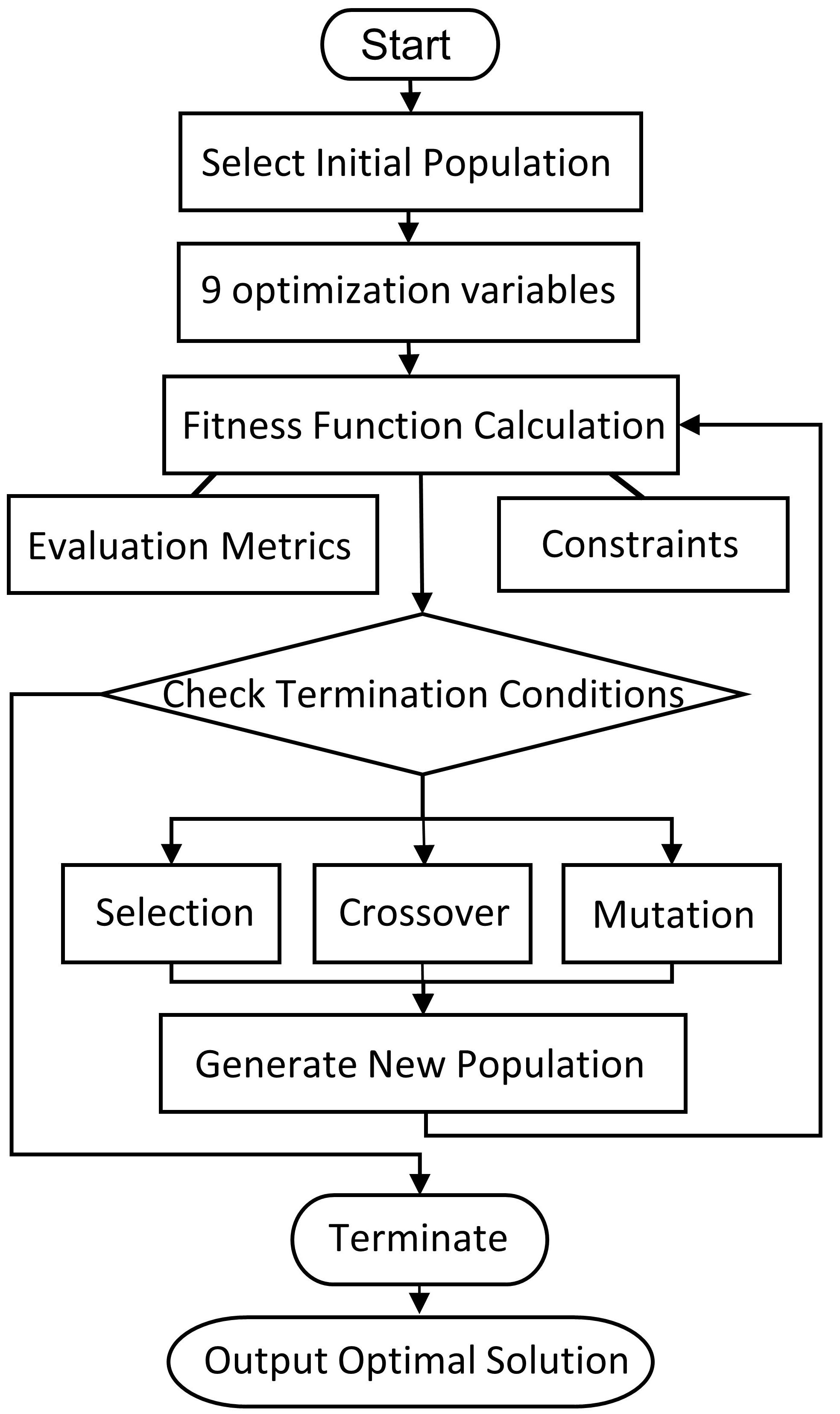}  
    \caption{Flowchart of GA-Based Parameter Optimization}
    \label{Fig.8}
\end{figure}

 In the iterative computation of a genetic algorithm, it is necessary to design an evaluation function, i.e., the fitness function, along with
corresponding constraints to ensure convergence. For the leg structure, the following constraints must be considered during the optimization process:

(1) Constraint on the farthest foot placement distance

Construction robots have specific obstacle-crossing requirements during locomotion. Therefore, when optimizing the leg dimensions, the influence of structural parameters on the farthest foot placement distance must be taken into account. A minimum value for the farthest foot placement distance is set as a constraint. Let the horizontal distance from the root joint axis to the farthest foot position along the planned trajectory be defined as the farthest foot placement distance \( D_m \). The constraint on the farthest foot placement distance is then expressed as:
\begin{equation}
\label{deqn_ex1}
D\geq(1-\lambda)D_0
\end{equation}
In the equation, \( D_0 \) represents the farthest foot placement distance for the initial dimensions; \( \lambda \) is the coefficient indicating the allowable reduction in the farthest foot placement distance after optimization, with \( \lambda = 0.05 \).

(2) Stiffness constraint:

Clearly, during the optimization process, reducing the cross-sectional dimensions of each leg segment can effectively decrease the mass and thereby lower the joint torques. However, this also leads to a reduction in deformation resistance. Therefore, it is necessary to constrain the stiffness of each leg segment during optimization. Since the legs are primarily subjected to bending moments during motion, the bending stiffness \( EI_z \) of each leg segment is defined. For the simplified hollow rectangular beam, the expression for \( EI_z \) is:
\begin{equation}
\label{deqn_ex1}
EI_z=E{\left[\frac{hw^3-(h-2t)(w-2t)^3}{12}\right]}
\end{equation}
During the optimization process,
\begin{equation}
\label{deqn_ex1}
EI_z\geq(1-\mu)EI_{z0}
\end{equation}

In the equation, \( t \) represents the wall thickness of the cross-section; \( \mu \) is the coefficient indicating the allowable reduction in stiffness after optimization, with \( \mu = 0.15 \); \( E \) is the elastic modulus of the material; and \( I_z \) is the moment of inertia of the cross-section.

The evaluation metrics for optimization are the joint torque and energy consumption, with equal weighting coefficients of 0.5 assigned to each. The objective is to minimize the combined objective function, which is constructed as follows:
\begin{equation}
\label{deqn_ex1}
\min F=0.5(\frac{T_{Mi}}{T_{Mi0}})+0.5(\frac{Q_i}{Q_{i0}})
\end{equation}

To solve constrained optimization problems, a penalty function can be introduced to construct an unconstrained optimization objective function. Therefore, incorporating the constraints defined above, a penalty function is added to the original fitness function to guide the population toward better evolution. The final fitness function is constructed as follows:
\begin{equation}
\label{deqn_ex1}
\begin{cases}
eval=0.5(\frac{T_{Mi}}{T_{Mi0}})+0.5(\frac{Q_i}{Q_{i0}})+k_p\times F_p \\
 F_p=F_T+F_Q+F_D+F_{EI} & 
\end{cases}
\end{equation}

In the equation, \( F_p \) is the penalty function, and \( k_p \) is the penalty coefficient, with \( k_p = 10000 \). The expression for \( F_p \) is as follows:
\begin{equation}
\label{deqn_ex1}
\begin{cases}
F_T=
\begin{cases}
0,\frac{T_{Mi}}{T_{Mi0}}\leq1 \\
\frac{T_{Mi}}{T_{Mi0}}-1 & 
\end{cases} \\
 \\
F_Q=
\begin{cases}
0,\frac{Q_i}{Q_{i0}}\leq1 \\
\frac{Q_i}{Q_{i0}}-1 & 
\end{cases} \\
 \\
F_D=
\begin{cases}
0,D\geq(1-\lambda)D_0 \\
(1-\lambda)D_0-D & 
\end{cases} \\
F_{EI}=
\begin{cases}
0,I_i\geq(1-\mu)I_{i0} \\
(1-\mu)I_{i0}-I_i & 
\end{cases} & 
\end{cases}
\end{equation}

Equation (40) is used as the final fitness function for the genetic algorithm, serving as a comprehensive metric for the peak joint torque and energy consumption of the leg. This function enables the achievement of a balanced optimal value for both metrics within the given constraints.
\subsection{Optimization Result Analysis}
In MATLAB, the initial population size is set to 100, with 100 iterations. The genetic algorithm is used to compute the defined evaluation function and constraints. The curve of the fitness function value against the number of iterations is shown in Fig.9. As can be seen from the figure, the fitness function value stabilizes after the 63rd iteration.
\begin{figure}[htb]
      \centering
      \includegraphics[width=\columnwidth]{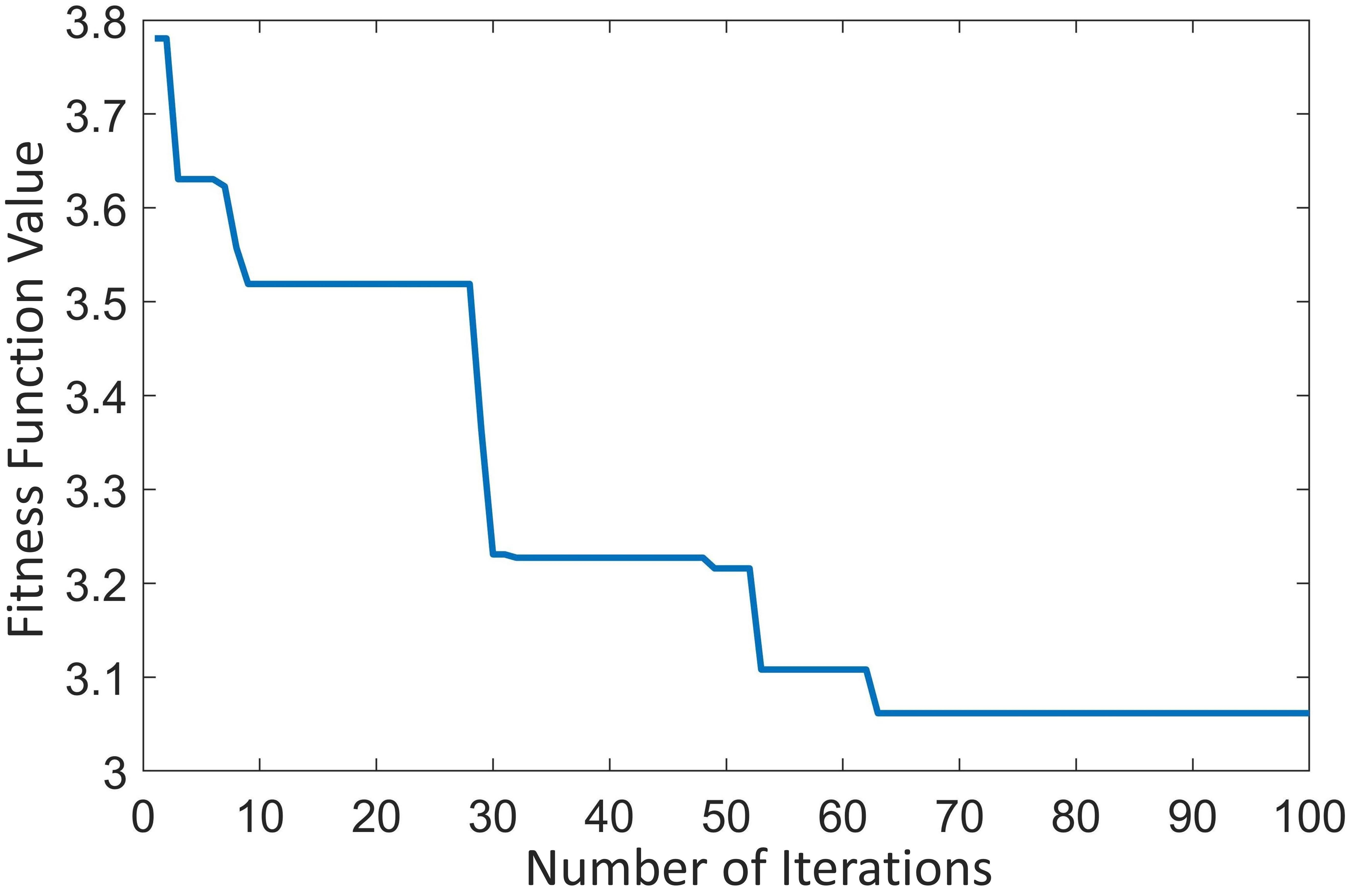}
      \caption{Fitness Function Variation Curve}
      \label{Fig.9}
   \end{figure}

After optimization using the genetic algorithm, the optimized structural parameters of the leg segments are obtained as shown in Table 2.
\begin{table}[ht]
    \centering
    \caption{Optimized Structural Parameters}
    \resizebox{8.5cm}{!}{%
      \begin{tabular}{cccc}
        \toprule
        \textbf{Leg Segment} & 
        \textbf{\makecell{Rod Length \\ $ l $ (mm)}} & 
        \textbf{\makecell{Cross-Sectional \\ Dimensions \\ $ h \times w $ (mm)}} & 
        \textbf{\makecell{Mass \\ $ m $ (kg)}} \\
        \midrule
        Coxa & 127 & $ 173 \times 115 $ & 5.20 \\
        Femur & 428 & $ 151 \times 169 $ & 17.17 \\
        Tibia & 446 & $ 130 \times 113 $ & 11.61 \\
        \bottomrule
      \end{tabular}%
    }
\end{table}

Based on the new structural parameters, the optimized peak joint torques are calculated, and a comparison of the peak joint torques before and after optimization is presented in Table 3.
\begin{table}[ht]
    \centering
    \caption{Comparison of Peak Torque Values Before and After Optimization for Each Joint}
    \resizebox{8.5cm}{!}{%
        \begin{tabular}{lccc}
            \toprule
            \textbf{Joint} & 
            \textbf{\makecell{Torque Before \\ Optimization (N$\cdot$m)}} & 
            \textbf{\makecell{Torque After \\ Optimization (N$\cdot$m)}} & 
            \textbf{\makecell{Reduction \\ Ratio (\%))}} \\
            \midrule
            Root Joint & 26.18 & 18.71 & 28.53 \\
            Hip Joint & 140.71 & 106.28 & 24.47 \\
            Knee Joint & 13.33 & 10.11 & 24.16 \\
            \bottomrule
        \end{tabular}%
    }
\end{table}

The joint torque curves before and after optimization are plotted in Fig.10. As can be seen from Table 3 and Fig. 10, the peak torques of all joints are significantly improved.
\begin{figure}[htb]
      \centering
      \includegraphics[width=\columnwidth]{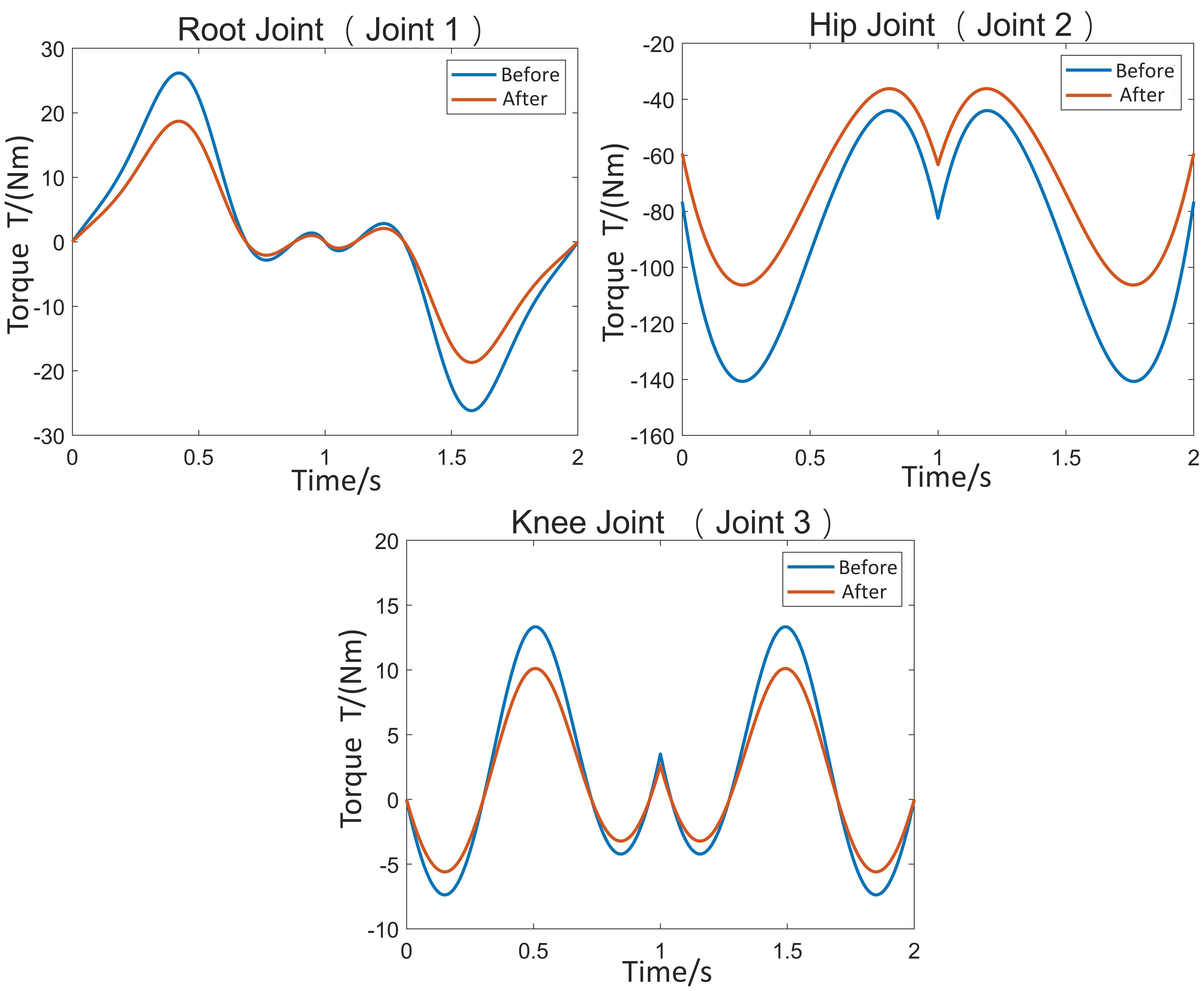}
      \caption{Torque Variation Curves of Each Joint Before and After Optimization}
      \label{Fig.10}
   \end{figure}

Using MATLAB, the total energy consumption of each joint after optimization is calculated according to Equation (35). The energy consumption results before and after optimization are listed in Table 4.

\begin{table}[ht]
    \centering
    \caption{Energy Consumption Comparison of Each Joint Before and After Optimization}
    \resizebox{8.5cm}{!}{%
        \begin{tabular}{lccc}
            \toprule
            \textbf{Joint} & 
            \textbf{\makecell{Consumption Before \\ Optimization (N$\cdot$m)}} & 
            \textbf{\makecell{Consumption After \\ Optimization (N$\cdot$m)}} & 
            \textbf{\makecell{Reduction \\ Ratio (\%))}} \\
            \midrule
            Root Joint & 11.31 & 8.09 & 28.47 \\
            Hip Joint & 292.60 & 226.47 & 22.60 \\
            Knee Joint & 22.33 & 16.94 & 24.14 \\
            \bottomrule
        \end{tabular}%
    }
\end{table}

As can be observed from the above figures and tables, the peak torques and energy consumption of all leg joints significantly decrease after optimization. The joint torques are reduced by approximately 20\%, and the energy consumption decreases by 24\% to 28\%, achieving the expected optimization objectives and validating the effectiveness of the optimization.

\section{DYNAMIC SIMULATION EXPERIMENT}
\subsection{Dynamic Simulation Model}

To further validate the rationality of the leg dimension optimization, dynamic simulation experiments are conducted on the leg before and after optimization. The ADAMS software enables realistic dynamic simulation analysis of the robot leg. The simulation model construction process is as follows: (1) The leg model created in SolidWorks is imported into ADAMS, and the material properties are defined. The main body of the leg is assigned 6061 aluminum alloy with a density of 2.77 g/cm³, and a global standard gravitational acceleration of 9.8066 m/s² is applied. (2) Single-degree-of-freedom revolute joints are added between the ground and coxa, coxa and femur, and femur and tibia to establish the kinematic relationships between the leg components. (3) The actuation of each leg joint is configured. Through these steps, the dynamic simulation model of the leg is completed, as shown in Fig. 11.

\begin{figure}[htb]
      \centering
      \includegraphics[width=\columnwidth]{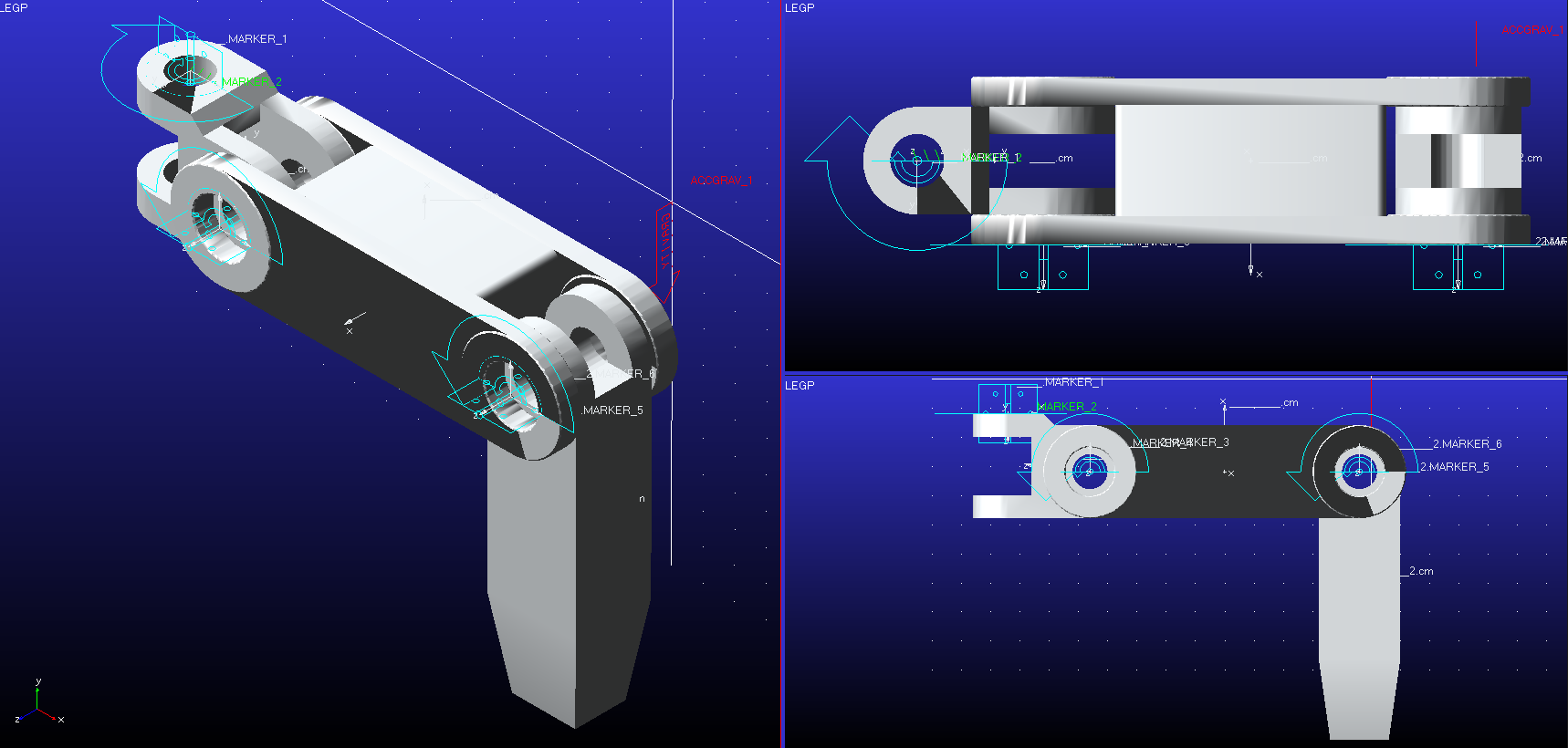}
      \caption{Dynamic Simulation Model of the Leg}
      \label{figurelabel}
   \end{figure}

Based on the optimization results from the previous section, the robot leg is reconstructed with the optimized dimensions, and a corresponding three-dimensional model is created and imported into ADAMS. Using the same method described above, the dynamic simulation model of the optimized leg is established.
\subsection{Simulation Experiment Data Processing}
First, an experiment is conducted on the first motion state of the leg before optimization. The time-varying functions of the joint angles from the starting point to the midpoint are imported into the Motion command settings for the revolute joints. The simulation step size is set to 100 steps, and the simulation duration matches the motion time defined in the trajectory planning. 
Through dynamic simulation analysis, ADAMS computes the power applied by the Motion drivers to each joint during movement. The effectiveness of the leg dimension optimization is verified by comparing the joint driving power before and after optimization. After performing the simulation based on the defined parameters, ADAMS outputs the time-varying curves of the joint driving power, as shown in Fig. 12.
\begin{figure}[htb]
      \centering
      \includegraphics[width=\columnwidth]{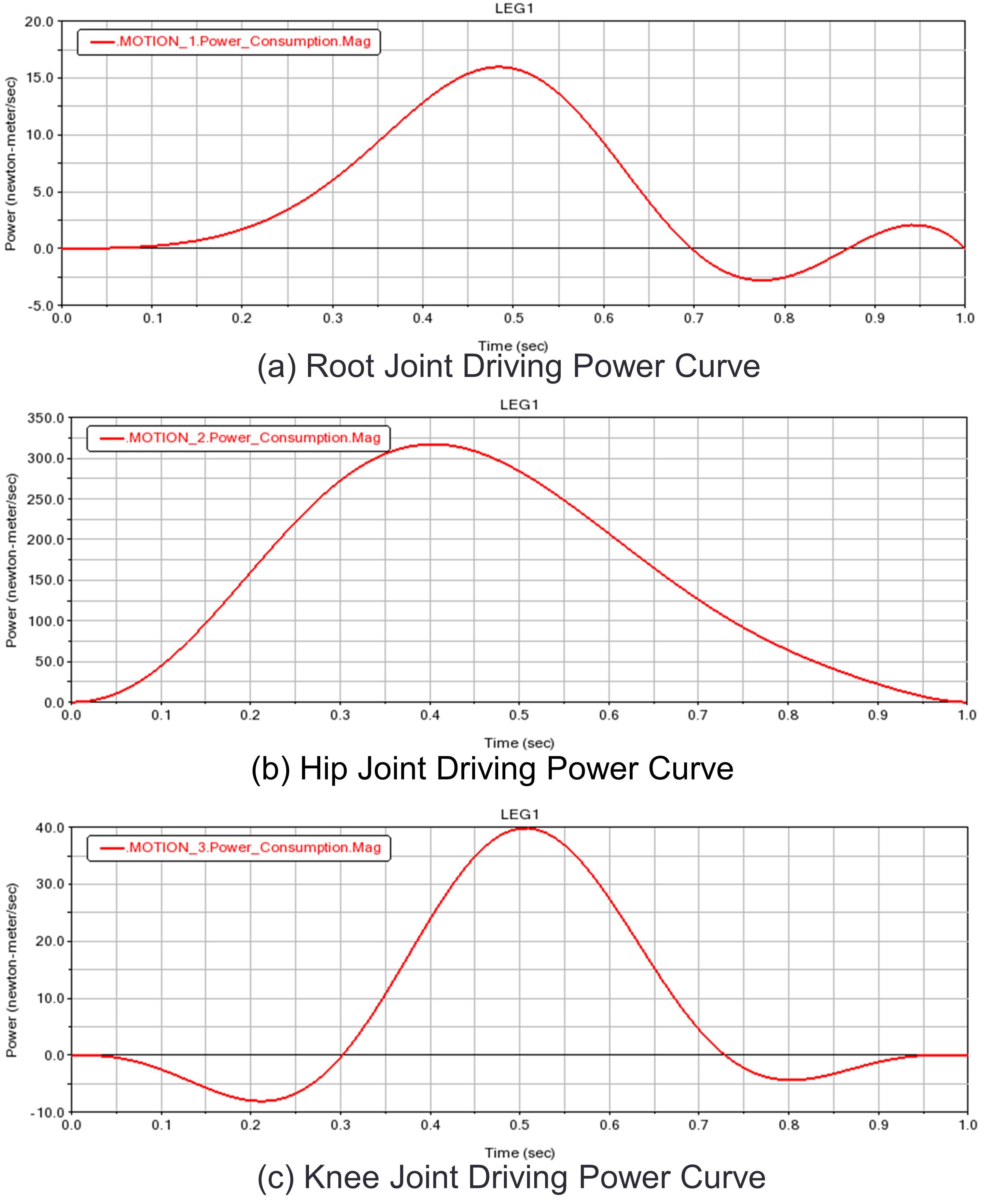}
      \caption{Joint Driving Power Curves Output by ADAMS}
      \label{figurelabel}
   \end{figure}

Simulations are conducted for two motion states of the leg before optimization: the movement from the starting point to the midpoint and from the midpoint to the endpoint. Corresponding simulations are also performed for the two motion states of the optimized leg. After computation, the output data from the software are exported and organized, and the driving power curves of each joint before and after optimization are plotted using MATLAB, as shown in Fig. 13.
\begin{figure}[htb]
      \centering
      \includegraphics[width=\columnwidth]{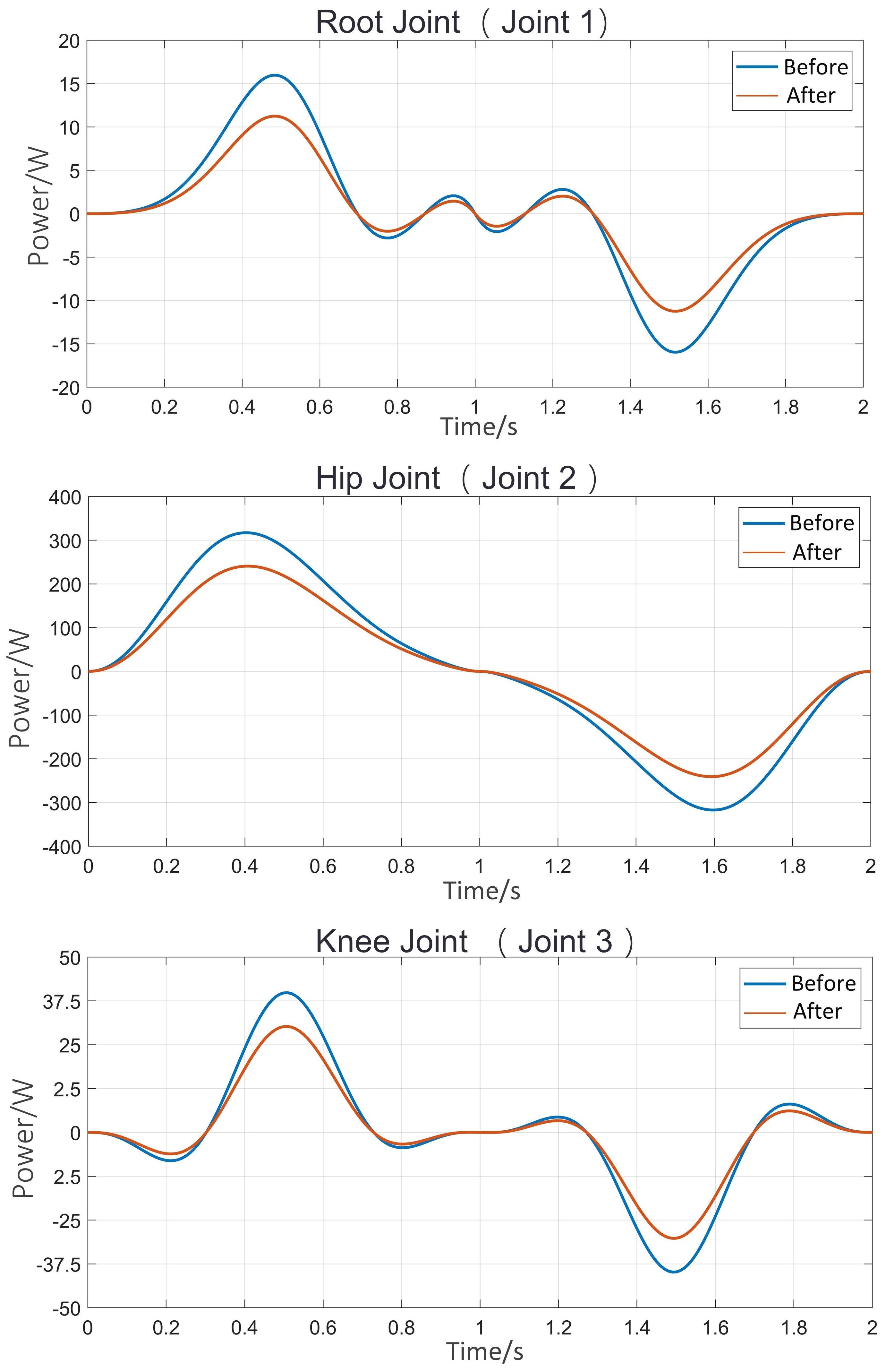}
      \caption{Driving Power Curves of Each Joint Before and After Optimization}
      \label{figurelabel}
   \end{figure}
As can be seen from the joint driving power curves obtained from the above simulations, both the overall and peak driving power of the optimized leg joints show a significant reduction compared to the pre-optimization values, with a decrease of approximately 20\%. This further demonstrates the validity of the optimization and confirms that the expected objectives have been achieved.

\section{CONCLUSIONS}

This study proposes a novel structural optimization method. By combining a dynamic model with the leg's motion trajectory, multiple dynamic evaluation metrics are formulated to comprehensively optimize the geometric size parameters of each leg segment. Through trajectory planning and dynamic modeling of the robot leg, the variations in joint torques are analyzed. Based on the dynamic model and motion trajectory, the peak joint torque and energy consumption are set as objectives, and a genetic algorithm is employed to iteratively solve for the leg's geometric size parameters, yielding the optimized dimensions. Experimental results demonstrate that the optimized leg structure achieves significant reductions in both peak torque and energy consumption, validating the effectiveness and rationality of the optimization strategy. Furthermore, dynamic simulation experiments conducted using ADAMS software further confirm the superior performance of the optimized leg in reducing driving power. In summary, this study provides a theoretical foundation and technical support for improving the dynamic performance of construction robot legs.

Despite the significant progress achieved in this study, there remains room for improvement in future research. First, in the process of dynamic modeling of the leg, the leg segments are treated as ideal rods with uniform mass and regular shapes to simplify calculations, while factors such as friction between joints and air resistance are not considered. Future work could focus on developing a more comprehensive dynamic model that accounts for additional real-world factors, enabling more accurate simulations of actual working conditions. Second, regarding the optimization method, although the genetic algorithm provides an effective solution, advancements in intelligent optimization algorithms continue to emerge. Exploring more advanced algorithms may offer new pathways to achieving more precise optimization results. This could not only further enhance the dynamic performance of construction robots but also broaden the application scope of related technologies, promoting innovation and development in construction robotics.

\addtolength{\textheight}{-8cm}   




\section*{ACKNOWLEDGMENT}
This work was supported by the National Key R\&D Program of China under Grant No.2023YFB4705002; National Natural Science Foundation of China under Grant No.U20A20283; Guangdong Provincial Key Laboratory of Construction Robotics and Intelligent Construction under Grant No.2022KSYS013; CAS Science and Technology Service Network Plan (STS) - Dongguan Special Project under Grant No.20211600200062; and the Science and Technology Cooperation Project of Chinese Academy of Sciences in Hubei Province Construction 2023.

This manuscript is currently under review at *Mechanics Based Design Of Structures And Machines*.

\end{document}